%% file: CHIL.tex
\documentclass[sigconf]{acmart}
\usepackage{booktabs}
\usepackage{color}
\usepackage{amsmath}
\usepackage{subcaption}
\usepackage{graphicx}
\usepackage{tikz}
\usepackage{xcolor}
\usepackage[labelfont=bf]{caption}
\usepackage[all]{nowidow}
\usepackage{longtable}

\AtBeginDocument{%
  \providecommand\BibTeX{{%
    \normalfont B\kern-0.5em{\scshape i\kern-0.25em b}\kern-0.8em\TeX}}}

\setcopyright{acmcopyright}
\copyrightyear{2020} 
\acmYear{2020} 
\setcopyright{acmlicensed}
\acmConference[ACM CHIL '20]{ACM Conference on Health, Inference, and Learning}{April 2--4, 2020}{Toronto, ON, Canada}
\acmBooktitle{ACM Conference on Health, Inference, and Learning (ACM CHIL '20), April 2--4, 2020, Toronto, ON, Canada}
\acmPrice{15.00}
\acmDOI{10.1145/3368555.3384462}
\acmISBN{978-1-4503-7046-2/20/04}
\usepackage{hyperref}  

\begin{document}

\title{Fast Learning-based Registration of Sparse 3D Clinical Images}

\author{Kathleen M. Lewis}
\email{kmlewis@mit.edu}
\affiliation{%
  \institution{Computer Science and Artificial Intelligence Laboratory (CSAIL), MIT}
}

\author{Natalia S. Rost}
\affiliation{%
  \institution{Massachusetts General Hospital, Harvard Medical School}}
  
\author{John Guttag}
\affiliation{%
  \institution{Computer Science and Artificial Intelligence Laboratory (CSAIL), MIT}}

\author{Adrian V. Dalca}
\affiliation{%
 \institution{Computer Science and Artificial Intelligence Laboratory (CSAIL), MIT \\ Massachusetts General Hospital, Harvard Medical School}}

\renewcommand{\shortauthors}{KM Lewis, et al.}

\begin{abstract}
We introduce SparseVM, a method that registers clinical-quality 3D MR scans both faster and more accurately than previously possible. Deformable alignment, or registration, of clinical scans is a fundamental task for many clinical neuroscience studies. However, most registration algorithms are designed for high-resolution research-quality scans. In contrast to research-quality scans, clinical scans are often sparse, missing up to 86\% of the slices available in research-quality scans. Existing methods for registering these sparse images are either inaccurate or extremely slow. We present a learning-based registration method, SparseVM, that is more accurate and orders of magnitude faster than the most accurate clinical registration methods. To our knowledge, it is the first method to use deep learning specifically tailored to registering clinical images. We demonstrate our method on a clinically-acquired MRI dataset of stroke patients and on a simulated sparse MRI dataset. Our code is available as part of the VoxelMorph package at \url{http://voxelmorph.mit.edu/}.
\end{abstract}


\maketitle

\section{Introduction}
For clinical neuroscience studies, accurate registration of clinical images is an important step in analysis \citep{stroke}. Unfortunately, existing registration algorithms are either too slow to be feasible for large population studies \citep{ants,patch} or are not sufficiently accurate on clinical images.

\begin{figure}[tb]
  \centering
  \includegraphics[clip, trim=0.25cm 3.25cm 0.25cm 3.25cm,width=0.5\textwidth]{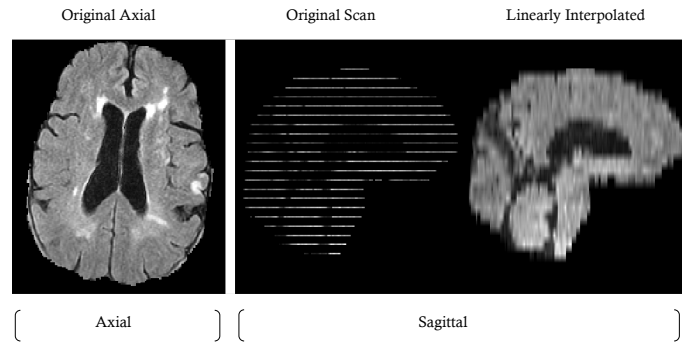}
  \caption{Example stroke subject scan with corresponding interpolated image. The subject scan has approximately 14\% of the slices normally available in a full resolution scan. Shown from left to right are an axial slice from the original scan, the original subject slices seen from a sagittal view, and a linearly interpolated sagittal image.}
  \Description{Example stroke subject scan with corresponding interpolated image. The subject scan has approximately 15\% of the slices normally available in a full resolution scan.}
  \label{volumes}
\end{figure}

Deformable medical image registration provides a dense, non-linear correspondence between a pair of medical scans. This correspondence, sometimes referred to as a flow field, can be used to warp one image to align with another image. Importantly the fields themselves are used in clinical studies to understand anatomical differences between scans. Existing registration methods usually work well on high-resolution research-quality scans. However, scans acquired in clinical settings can be problematic for these methods. Clinical settings often require limited scanning time because of patient safety and financial constraints. For 3D imaging modalities such as structural MRI, this often means only a few 2D slices are acquired instead of a dense set of slices, leading to spatially sparse scans. Each 2D slice has high in-plane resolution (Fig. \ref{volumes}, left), but the 3D volume can be missing up to 86\% of the slices that are typically available in a full resolution scan (Fig. \ref{volumes}, right). While slice thickness tends to be higher in these scans as well, the often much higher slice separation is the predominant issue. The wide spacing between slices causes drastic discontinuities in anatomy between neighboring slices, and these discontinuities lead to a reduction in the registration accuracy.

In this paper, we present a new learning-based registration method, SparseVM, that adapts the recent VoxelMorph method \citep{VM,VMMICCAI,VMArxiv} to clinical images. To our knowledge, this is the first use of deep learning based strategies specifically designed for registering sparse clinical images. We evaluate SparseVM on a neuroimaging study of stroke patients containing T2-FLAIR MR brain scans \citep{stroke} and on a T1 MR brain dataset from the ADNI dataset \citep{ADNI} with simulated sparsity. 

Through qualitative and quantitative analysis, we show that SparseVM provides significant improvements in both speed and accuracy over the current state-of-the-art methods. Specifically, SparseVM is 1) more accurate and 1000$\times$ faster than the best classical registration methods \citep{ants,patch} while also having a higher accuracy, and 2) more accurate than current best learning-based methods \citep{VM}. 

%SparseVM outperforms the fastest baseline and state-of-the-art learning-based method \citep{VM} on 91\% of test subjects from the stroke dataset and on 81\% of test subjects from the simulated dataset. Additionally, we provide quantitative and qualitative analysis of our design decisions and their effect on registration performance. 

\section{Related Work}

%\begin{figure}[t!]
 % \centering
 % \includegraphics[width=\textwidth]{mlhc/VoxelMorphvsPBR.png}
 % \caption{Learning-based methods are much faster than iterative optimization methods for the clinical image registration task, but perform less accurately.}
 % \label{PBRvVM}
%\end{figure}

Most classical registration methods are high-dimensional optimization algorithms designed for full-resolution scans rather than sparse clinical scans. Methods include elastic-type models \citep{elastic, elastic1,elastic2}, statistical parametric mapping \citep{stats}, free-form deformations with b-splines \citep{freeform}, discrete methods \citep{mrf}, and Demons \citep{demons, demons1}. 
Diffeomorphic transforms explicitly enforce topology constraints  \citep{geodesic, diffMedial,DARTEL,diffDemons,CC}. Applying these methods to clinical scans is problematic because they were developed to work on high resolution scans and require spatial continuity for smooth image gradients, which are not available in clinical scans. Additionally, these methods are extremely slow because they require solving an iterative optimization problem for each new pair of images.

More recently, patch-based registration (PBR) \citep{patch} builds on previous discrete registration methods \citep{mrf} and adapts them to sparse data by using sparse 3-dimensional patches to measure image intensity similarities. PBR incorporates a mask that weights voxel contributions to the loss function in proportion to confidence in those voxels. This differentiates interpolated voxels from the voxels acquired from actual slices. PBR is a flexible, accurate method for registering clinical images, but it employs a pairwise optimization strategy, requiring on the order of two CPU hours to register a pair of images. This means that using it to register a large database of clinical images is very time-consuming.

Recently developed learning-based methods can register dense scans significantly faster than optimization-based registration methods. They use neural networks to learn a function that takes in two scans as input and outputs a deformation field. Many are supervised, requiring ground truth warp fields during training \citep{shape, nonrigid, quicksilver, steer}. Unsupervised methods \citep{endtoend,fan,BIRNet,Faim} have been shown to work well on research quality, high resolution scans. VoxelMorph (VM) \citep{VM,VMArxiv, VMMICCAI}, a recent unsupervised learning-based method, achieves state-of-the-art accuracy on 3D isotropic scans and is fast at test time. However, on clinical images, VoxelMorph does not achieve the same accuracy as PBR, as we illustrate in our experiments.

In this paper, we present a method that combines the fast runtimes of learning-based methods with insights derived from classical methods for sparse data, to achieve both high performance and state-of-the-art accuracy. Specifically, we design a new loss function, Sparse Local Cross Correlation (SLCC), for deep learning based registration that weights acquired voxels differently from interpolated voxels, and we analyze the hyper-parameters for this function. We demonstrate that this achieves the fast runtimes of learning-based methods and exceeds the accuracy of both classical and learning-based methods.

\section{Method}

We build on the learning-based method VoxelMorph to register clinical scans. VoxelMorph uses a convolutional neural network to learn a function, $g_\theta(f,m) = \phi$, where $\theta$ are network parameters, that computes a deformation field $\phi$ for a pair of scans \{$f,m$\}. At test time, VoxelMorph evaluates this function on two input scans and outputs the deformation field $\phi$, as well as the registered image, $m \circ \phi$ ($m$ warped by $\phi$). 

The essential difference between our method and VoxelMorph is that we introduce and incorporate a novel loss function, sparse local cross correlation (SLCC), designed to compensate for the sparseness in the image. SLCC, described in detail below, uses two masks and weights the contributions of each input 3D image in proportion to values in the respective masks. This loss function can therefore be used in any application where sparsity (binary values) or confidence in observations (continuous values) can be defined via masks.

%\textcolor{red}{Moved from introduction ; find if need to put it somewhere ... We introduce a new loss function, sparse local cross correlation (SLCC), that generalizes beyond the clinical registration of MR brain scans shown in this paper. SLCC takes in two masks and weights the contributions of each input in proportion to values in the respective masks. This loss function can therefore be used for in any application where sparsity (binary values) or confidence in observations (continuous values) can be defined by masks.}

\subsection{Loss Functions}
In this section, we first describe the loss function used in VoxelMorph. 
Let \textit{f}, \textit{m} be two volumes defined over a 3D spatial domain $\Omega \subset \mathbb{R}^{3} $. We assume \textit{f} and \textit{m} contain grayscale data and are affinely aligned (we do this using one of the publicly available affine alignment packages \citep{freesurfer}) as a preprocessing step, so that the scan pairs exhibit only non-rigid misalignments. 

Image registration methods optimize a loss function of the form:
\begin{gather}\label{VMloss}
    L(f,m,\phi) = L_{sim}(f, m\circ\phi) + \lambda L_{smooth}(\phi),
\end{gather}
where $\lambda$ is a regularization hyperparameter. This loss function balances an image matching term $L_{sim}$ with a regularization or spatial smoothness term $L_{smooth}$. The first term measures the similarity in appearance between the fixed image $f$ and the warped moving image $m\circ\phi$. The second term encourages the displacement field $\phi$ to be smooth and anatomically realistic. This loss is used in classical methods to optimize each deformation independently and in unsupervised learning-based methods to learn the global network parameter $\theta$.

The $L_{smooth}$ term encourages neighboring voxels to move together. Without this loss, the optimization can lead to an anatomically meaningless deformation field that moves voxels from arbitrary locations to match the intensities of the fixed image. Specifically, the $L_{smooth}$ term penalizes the local spatial gradients of $u$, $\bigtriangledown u$, where $u$ is the vector displacement such that $\phi = Id + u$ and $Id$ is the identity transform:
\begin{gather}\label{gradientloss}
L_{smooth}(\phi) = \sum_{p \in \Omega} || \bigtriangledown u(p)||^2,
\end{gather}
where the notation $u(p)$ indicates the value of $u$ at voxel location $p$.

Intensity mean-squared error (MSE) and cross correlation (CC) are the most commonly used similarity measures, $L_{sim}$. MSE compares intensities at each voxel:
\begin{gather}\label{eq:VMMSE}
    MSE(f, m\circ\phi) = \frac{1}{|\Omega|}\sum_{p \in \Omega}[f(p) - [m\circ\phi](p)]^2.
\end{gather}
This measure works well when the dataset contains images with similar intensity distributions without intensity anomalies or excessive noise. 

Local normalized cross correlation has been shown to be more invariant than MSE to small differences in intensity distributions and noise across scans \citep{CC}. For a scan $f$, let $\mu_f(p)$ be the local neighborhood intensity mean around voxel $p$: $\mu_f(p) = \frac{1}{n^3}\sum_{p_i}f(p_i)$, where $n^3$ is the neighborhood size and $p_i$ iterates over the $n^3$ volume of voxels surrounding voxel p. Let the local mean subtracted from a voxel in its neighborhood be defined as: $\bar{f}(p, p_i)= f(p_i) - \mu_f(p)$ and similarly $\bar{m}_\phi(p,p_i)= [m\circ\phi](p_i) - \mu_{m\circ\phi}(p)$.  The local normalized cross correlation is then:
\begin{gather}\label{eq:VMCC}
    LCC(f, m\circ\phi) = \sum_{p \in \Omega}\frac{[\sum_{p_i}\bar{f}(p,p_i) \bar{m}_\phi(p,p_i)]^2}{[\sum_{p_i} \bar{f}(p,p_i)^2][\sum_{p_i} \bar{m}_\phi(p,p_i)^2]}.
\end{gather}
\subsection{Sparse Method}\label{sparsemethods}
We introduce two new image similarity losses that generalize Eq. \eqref{eq:VMMSE}
and Eq. \eqref{eq:VMCC} and lead to accurate registration on both clinical and high resolution scans. We start by up-sampling each clinical image to isotropic spatial resolution (equal resolution in all planes) by linearly interpolating between the acquired slices (Fig. \ref{volumes}, right).

We define masks, $w_f$ and $w_m$, corresponding to the fixed image $f$ and the moving image $m$, respectively. In its simplest form, each voxel in a mask has a value of 0 (interpolated voxel) or 1 (acquired voxel). We explain other forms in the experiments section. We let $w_c = w_f * w_m$ be a combined mask that indicates voxels that are observed in both scans, where $*$ denotes element-wise multiplication. Similar masks were first used in \citep{patch} to define patch-wise squared error similarities.

Sparse mean squared error (SMSE) generalizes Eq. \eqref{eq:VMMSE} by computing the loss over only observed voxels:
\begin{gather} \label{eq:sparse_mse}
SMSE(f, w_c, m\circ\phi) = \frac{1}{\sum w_c(p)}\sum_{p \in \Omega} w_c(p)[f(p) - [m\circ\phi](p)]^2.
\end{gather}

To define a sparse version of local normalized cross correlation, we first consider the sparsity of each local neighborhood of voxels. For each image, we compute the local mean of each neighborhood over only the observed voxels, and then subtract the local mean from each pixel in the neighborhood:
\begin{gather}\label{eq:sparse_CC_mean}
    \begin{aligned}
&\mu_f^w(p) = \frac{1}{\sum_{p_i \in n^3}w_f(p_i)}\sum_{p_i \in n^3}w_f(p_i)f(p_i),
    \end{aligned}
\end{gather}
\begin{gather}
    \begin{aligned}
&\bar{f}(p,p_i) = f(p_i) - \mu_f^w(p)    
    \end{aligned}
\end{gather}
with similar definitions for $m\circ\phi$.
%We next define $\hat{f}(p_i,p)$  at voxel $p_i$ belonging to a neighborhood centered at $p$ as the image with the local mean subtracted out:  
% 
%\begin{gather}\label{eq:sparse_CC_num}
%    \begin{aligned}
 %   &\hat{f}(p_i,p) = f(p_i) - \mu_f(p).
%    \end{aligned}
%\end{gather}
%  
Finally we define the sparse local cross correlation as: 
\begin{gather}\label{eq:sparse_CC}
    \begin{aligned}
    &SLCC(f, w_c, [m\circ\phi]) = 
    \\
    &\sum_{p \in \Omega}\frac{[\sum_{p_i \in n^3}w_c(p_i)\bar{f}(p,p_i)\bar{m}_\phi(p, p_i) ]^2}{[\sum_{p_i \in n^3}w_c(p_i)\bar{f}(p,p_i)^2][\sum_{p_i \in n^3} w_c(p_i)\bar{m}_\phi(p,p_i)^2]}.
    \end{aligned}
\end{gather}
% 
%\begin{gather}\label{eq:sparse_CC}
%    \begin{aligned}
%    &SLCC(f, w_c, [m\circ\phi]) = 
%    \\
%    &\sum_{p \in \Omega}\frac{[\sum_{p_i \in n^3}w_c(p_i)(f(p_i) - \hat{f}(p))([m\circ\phi](p_i) -\hat{k}(p) )]^2}{[\sum_{p_i \in n^3}w_c(p_i)(f(p_i) - \hat{f}(p))^2][\sum_{p_i \in n^3} w_c(p_i)([m\circ\phi](p_i) -\hat{k}(p) )^2]},
 %   \end{aligned}
%%\end{gather}
%

SLCC computes the local correlation in each neighborhood between voxels that are observed in both scans. Since SLCC does not use values over unobserved (interpolated) voxels, the neighborhoods contain sparse observations. As in regular local normalized cross correlation, the neighborhood size $n^3$ is an important hyperparameter. In SLCC, special care must be taken to ensure the neighborhoods are large enough to contain enough non-zero voxels, but small enough to capture statistics of specific regions of the scans. In our experiments, we provide an analysis of this hyperparameter.

The smoothness constraint on the deformation field Eq. (\ref{gradientloss}) remains the same regardless of whether the data is sparse or not, since we still want to regularize the underlying deformation field to be smooth and anatomically realistic at all voxel locations.

\section{Experiments}

\begin{table*}
\begin{center}
\begin{tabular}{l|c|c|c|c}
    \textbf{Method} & \textbf{Average (SD)} & \textbf{Median (MAD)} & \textbf{CPU sec} & \textbf{GPU sec}\\
    \hline
    ANTs (optimized) & 0.696* (0.070) & 0.720 (0.050) &  9059* (2023) &  - \\
    PBR (optimized)  & 0.732 (0.071) & 0.749 (0.052) & 9269* (5134)  & -  \\
    VoxelMorph LCC & 0.723* (0.077) & 0.748 (0.060) & 9.05 (0.13) & 0.27 (0.05) \\
    SparseVM SMSE & 0.719* (0.075) & 0.740 (0.058) & 9.04 (0.13) & 0.27 (0.05) \\
    SparseVM SLCC & \textbf{0.743 (0.076)} & \textbf{0.768 (0.060)} & 9.10 (0.15) & 0.28 (0.07)  \\
    \hline
\end{tabular}
\end{center}
\caption{ Dice scores and test runtimes for ANTs, PBR, VoxelMorph, and SparseVM on the stroke dataset test subjects. The average Dice score (standard deviation in parentheses) and median Dice score (median absolute deviation in parentheses) are computed for all test subjects on the left and right lateral ventricle labels. The runtimes are computed after pre-processing, which is the same for all methods. We use a * symbol to indicate statistical significance between sparseVM and a baseline, using a paired t-test with a 0.01 threshold. SparseVM using SLCC is more accurate than all baselines while being as fast or faster.}
\label{tab:onecol}
\end{table*}

\begin{figure*}
    \centering
    \includegraphics[width=\textwidth]{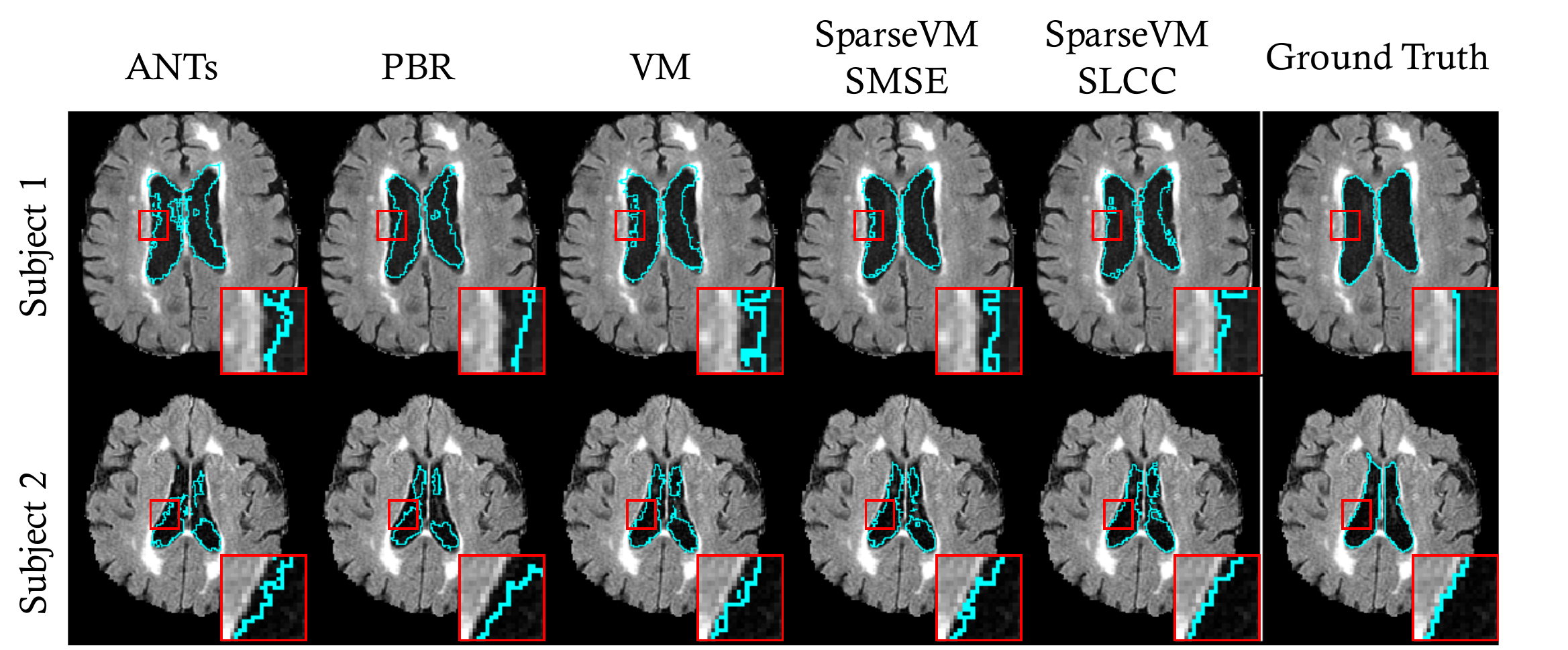}
    \caption{Example stroke scan results from the baselines and SparseVM. Warped atlas segmentations of the ventricles are overlaid in blue on a test subject scan for each method and the ground truth, with close-up regions illustrating the registration accuracy near ventricle boundaries.}
    \label{fig:example_segs}
\end{figure*}

\begin{figure}[t!]
  \centering
  \includegraphics[width=0.5\textwidth]{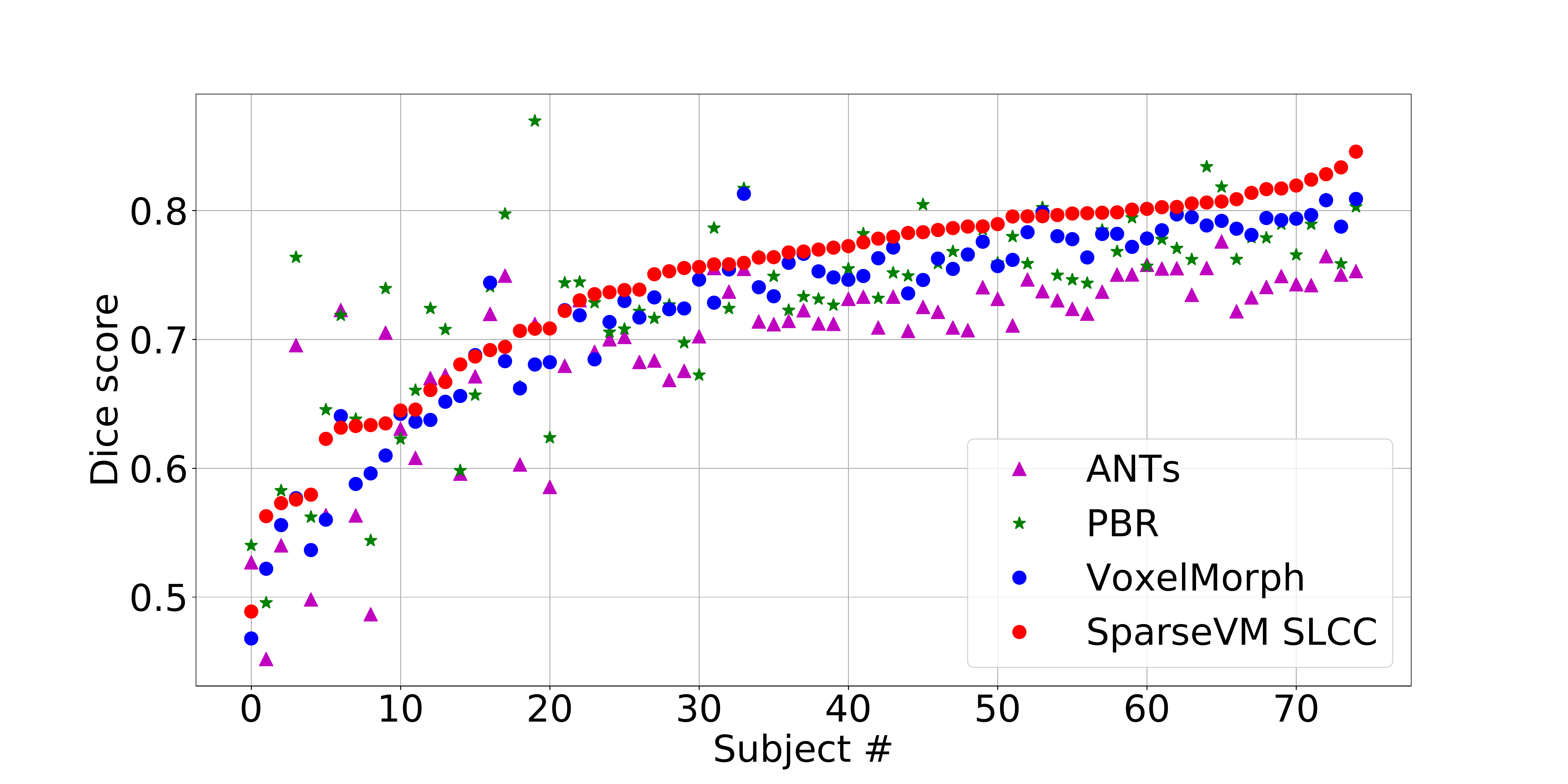}
  \caption{Dice score (higher is better) for each scan sorted by SparseVM performance. SparseVM using SLCC yields better Dice scores than ANTs on 86.7\% of test subjects, better than PBR on 69.3\% of test subjects, and better than VoxelMorph on 90.7\% of test subjects.}
  \label{dotplot}
\end{figure}

\begin{figure}[]
    \centering
    \includegraphics[width=0.5\textwidth]{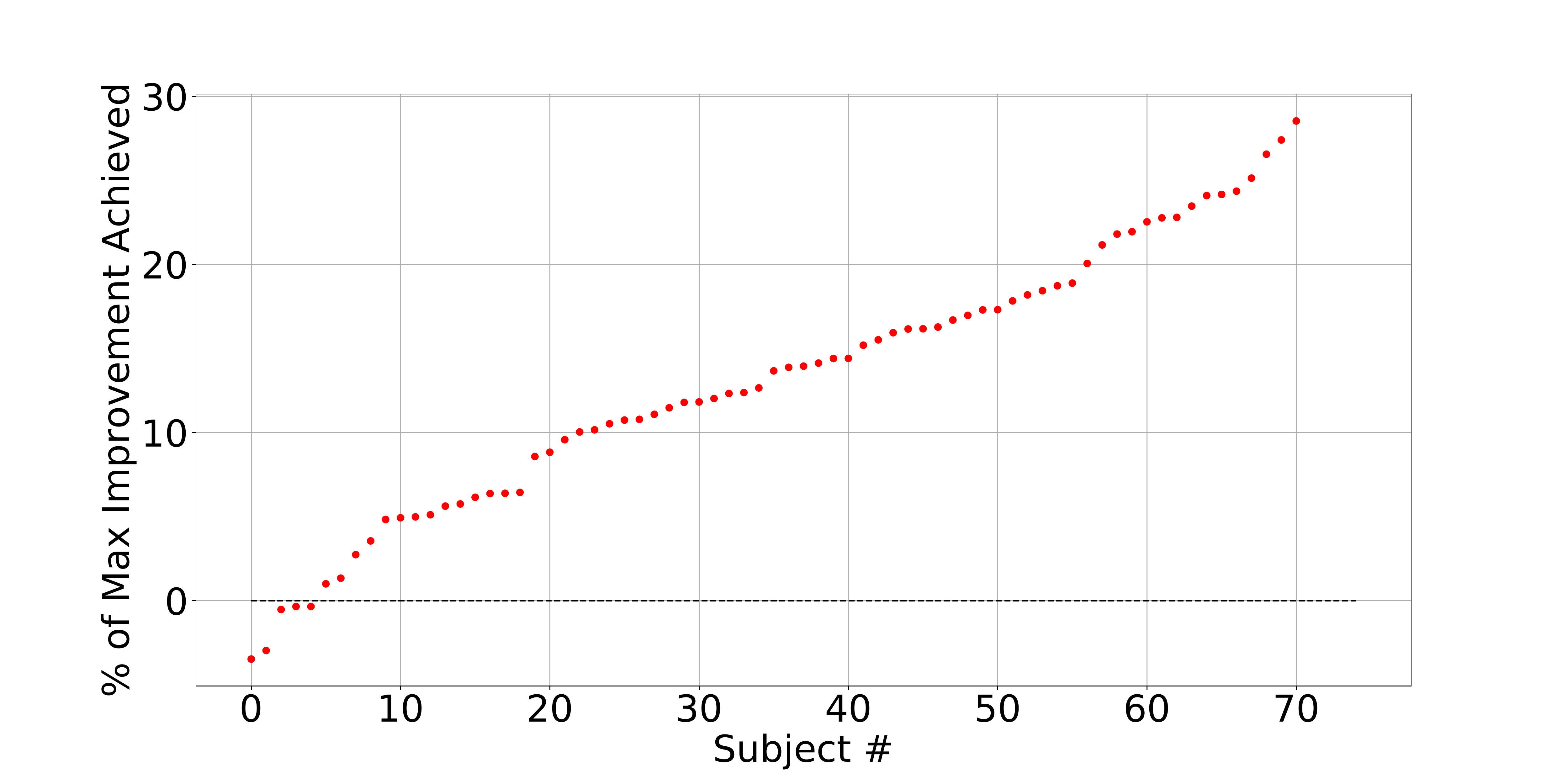}
    \caption{Improvement of SparseVM over Voxelmorph as a percentage of maximum achievable difference. The data are sorted in increasing order of percentage. There are four outliers not shown. For two of the outliers, SparseVM obtains $>40\%$ of the possible improvement over VoxelMorph in terms of Dice. For the other two outliers, SparseVM is much worse than VM. Upon visual inspection, we find that these two scans were incorrectly skull-stripped with a significant part of the brain removed.}. 
    
    \label{fig:dice_stroke_improvement}
\end{figure}

\begin{figure*}[tb!]
\centering
\includegraphics[width=0.8\textwidth]{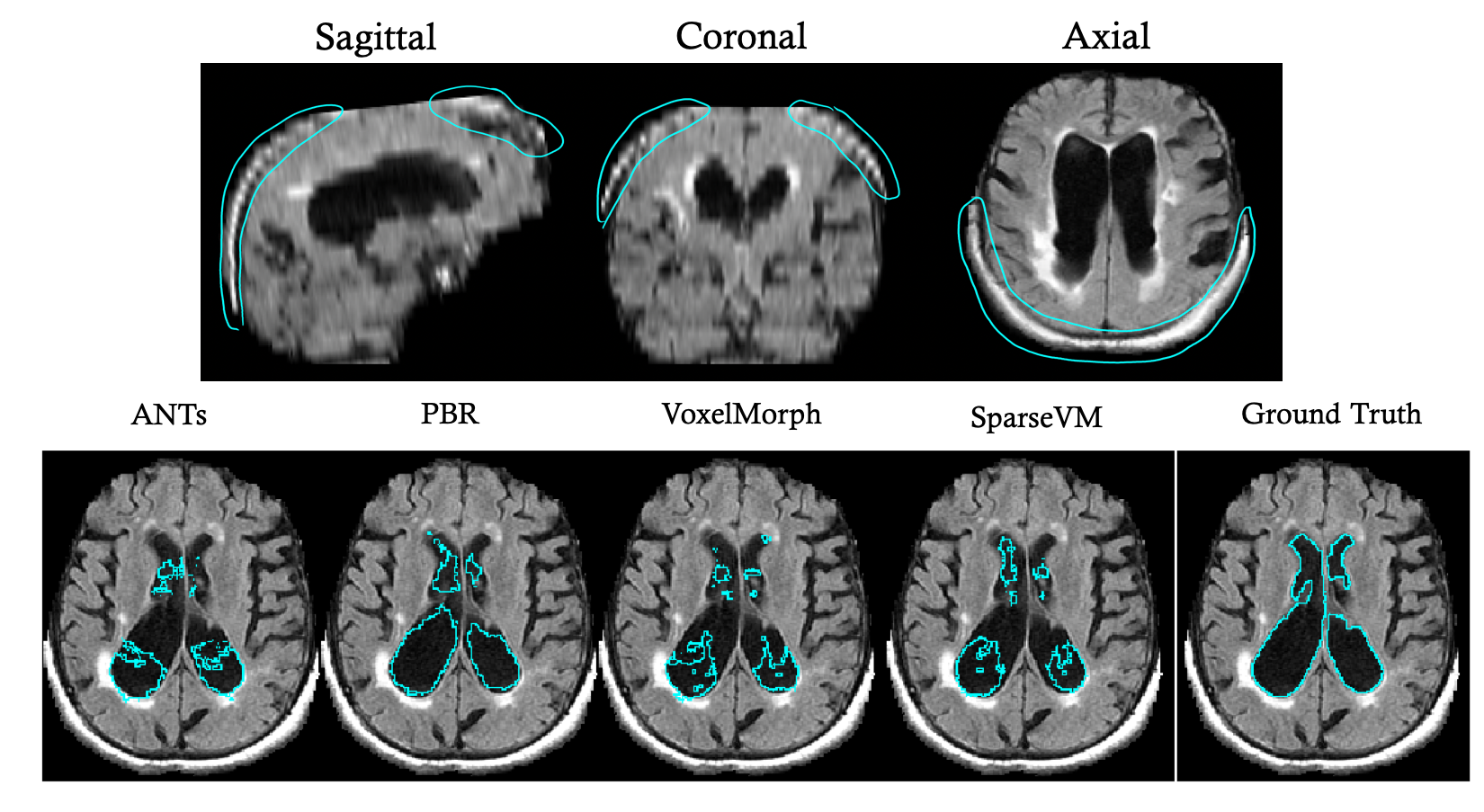}
\caption{Test subject that was not properly skull stripped during pre-processing. The top row highlights examples of areas where skull is present in the scan. The bottom row shows the warped atlas ventricle segmentations for each method. The PBR segmentations align well with the ventricles, while the other methods only segment part of the ventricles.}
\label{bad_vol_SparseVM}
\end{figure*}

\begin{figure*}[tb!]
  \centering
  \includegraphics[width=\textwidth]{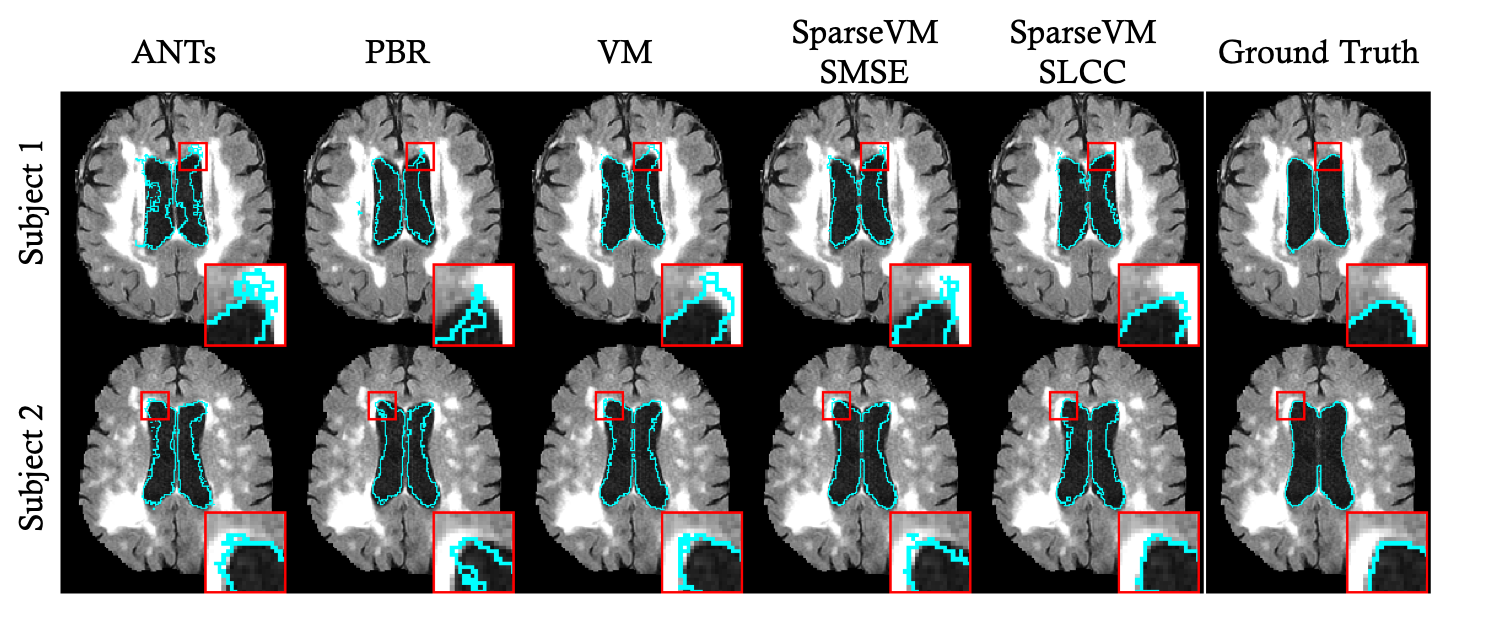}
  \caption{Example stroke scan results for subjects with high white matter hyperintensity (WMH). Warped atlas segmentations of the ventricles are overlaid in blue on a test subject scan for each method and the ground truth, with close-up regions illustrating the registration accuracy near ventricle boundaries.}
  \label{WMH_volumes}
\end{figure*}

\begin{figure*}[t]
  \centering
  \includegraphics[clip, trim=3cm 0cm 4cm 1.5cm,width=0.8\textwidth]{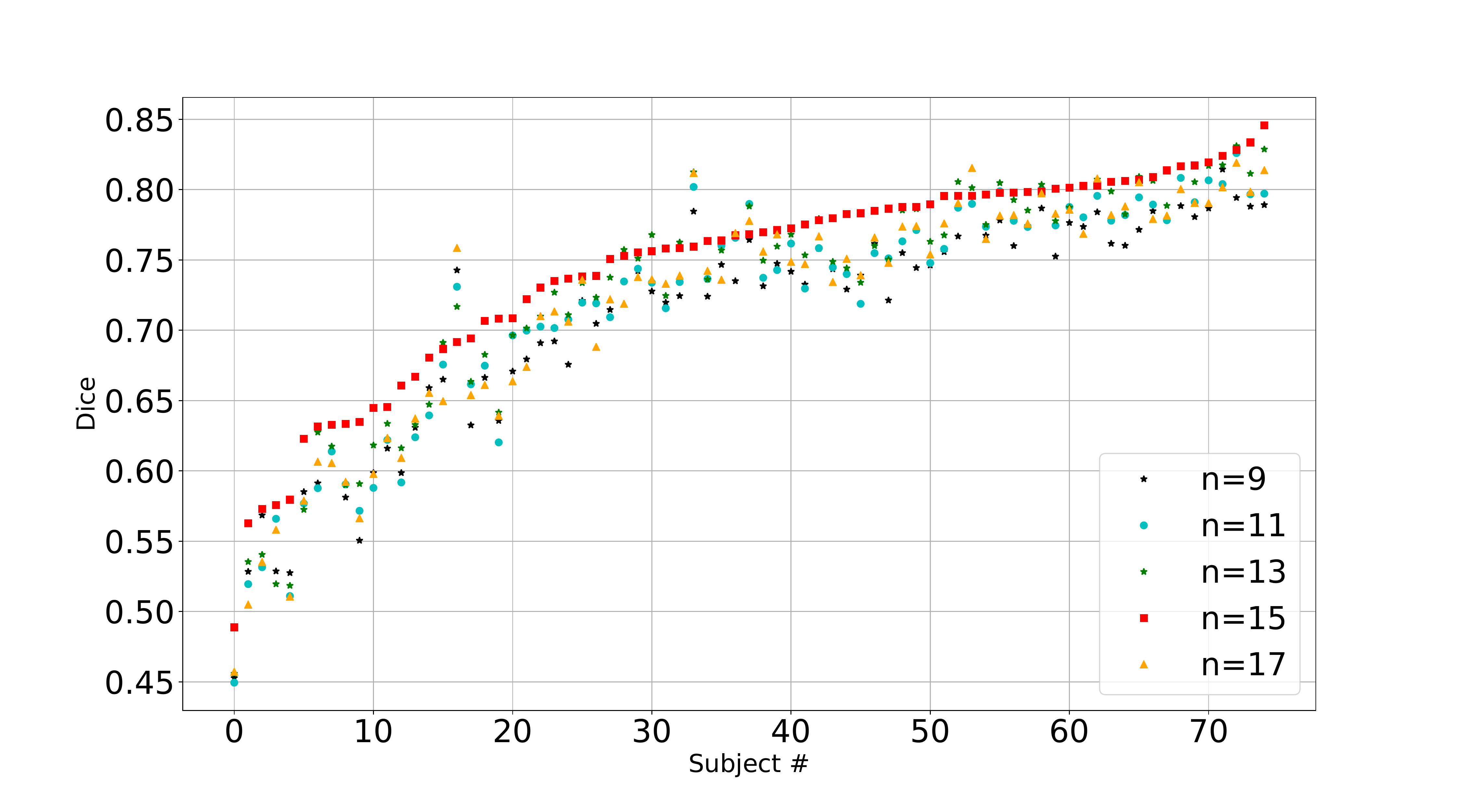}
  \caption{Dice scores for registered subjects using SparseVM SLCC models with different neighborhood parameter $n$. The Dice scores are sorted in increasing order of $n=15$ Dice scores. The graph shows that $n=15$ is the parameter that results in the highest Dice scores for SparseVM using SLCC.}
  \label{neighborhood}
\end{figure*}

In our experiments, we focus on atlas-based registration, registering an atlas, or reference scan, to each subject scan. This is often done in population studies, where inter-subject registration is a core problem. We demonstrate our method on a clinically acquired MRI dataset for stroke patients \citep{stroke} and on a simulated sparse dataset derived from the ADNI dataset \citep{ADNI}. We use full-resolution atlases computed for each modality \citep{atlasstudy}. 

%which contains manual segmentations of the ventricles.

\subsection{Datasets}
The clinically-acquired dataset contains T2-FLAIR MR scans acquired within 48 hours of stroke onset. The in-plane resolution is 0.85mm with varying slice thicknesses and the spacing between slices ranges from 5-7mm. The dataset contains manual segmentations of the ventricles for 104 of the subjects. The 237 subjects without manual segmentations were used for unsupervised registration training, while the 104 subjects with manual segmentations were split into validation (29 subjects) and held out test (75 subjects) sets. 

The second dataset contains isotropic 1mm T1-weighted brain scans from the publicly available Alzheimer’s Disease Neuroimaging Initiative (ADNI) dataset \citep{ADNI}. This dataset has the advantages of having segmentations of 30 different structures in the brain obtained using FreeSurfer \citep{freesurfer}. We use this dataset to create simulated clinical scans as follows:

\begin{enumerate}
    \item Simulating the variance in head positioning during acquisition by applying small, random affine changes to the isotropic image before creating sparse data.
    \item Deleting 6 out of every 7 axial slices. The resulting images contain approximately 14\% of the acquired voxels in the isotropic image. The number of slices deleted was chosen to emulate the wide spacing in the clinically acquired stroke dataset.
    \item Simulating motion blur by combining, in the frequency domain, a simulated sparse image with a slightly shifted version of itself.
\end{enumerate}

These steps attempt to realistically mimic challenges that occur during clinical image acquisition. This enables more detailed analyses on a variety of anatomical regions than is possible with the clinical dataset. These images were affinely registered to a T1-weighted atlas.

During pre-processing the scans for both datasets were intensity normalized, affinely-aligned, and skull-stripped using existing toolboxes \citep{atlasstudy,ants,Robex}.

\subsection{Masks}
The atlas $m$ in each atlas-based registration experiment is isotropic, and therefore $w_m$ consists of the value 1 at each voxel. The clinical (or simulated clinical) images $f$ have approximately 14\% of the number of acquired slices of an isotropic research scan. We linearly interpolate the remaining slices to achieve the same resolution as the atlas. The subject mask $w_f(p)$ is set to 1 for voxels in the acquired slices and zero for voxels in the interpolated slices. While performing affine alignment, both $f$ and $w_f$ are resampled, and voxels are therefore linearly interpolated. The masks are also linearly interpolated, which means the final mask values in $w_f$ are continuous in the range $[0,1]$. We interpret these values as weightings of confidence in the values of the voxels.

\subsection{Baseline Methods}
For the clinically-acquired stroke dataset, we compare our method to VoxelMorph (VM) \citep{VM}, Patch Based Registration (PBR) \citep{patch}, and the widely used Advanced Normalization Tools software package (ANTs) \citep{ants}. 

For VM and PBR, we used the optimal parameters given in the original publications, except for the cross correlation neighborhood size. In the original VoxelMorph paper, a cross correlation neighborhood size of $9^3$ is used. However, from our experiments we found a larger neighborhood size of $15^3$ leads to better performance for clinical images, so we report results with a neighborhood size of $15^3$. This is likely because of the sparsity of the neighborhoods in clinical data. ANTs is used as a baseline in the original VoxelMorph and PBR papers. For ANTs, we use the cross correlation loss, and parameters optimized for Dice score: (SyN step size of 0.25 and Gaussian parameters (9,0.2) at three scales with a maximum of 201 iterations per scale).

\subsection{Evaluation Metric}
We use the output deformation field to warp segmentation maps of the atlas to the subject and compare to the ground truth subject segmentations using the Dice score \citep{dice}. We compute the Dice score for acquired slices in the clinical scan.

The Dice score measures the volume overlap of anatomical segmentations and ranges between 0 and 1, with 0 representing no overlap and 1 representing perfect overlap. However, when segmentations are manually annotated by humans, there can be considerable inter-rater variability. This often leads to lower Dice scores even for segmentation tasks. Furthermore, state-of-the-art registration studies on high-quality, high-resolution MRI modalities have reported at most a ventricle average Dice score of 0.9 \citep{VM,klein}.

\subsection{Implementation}\label{sec:implementation}
We implemented SparseVM using Keras \citep{keras} with a TensorFlow backend \citep{tf}. We used the ADAM optimizer \citep{adam} with an $\epsilon$ value of $5 \times 10^{-5}$ and a learning rate of $5 \times 10^{-4}$. We used a modified version of the U-Net-like \citep{Unet} architecture proposed in the original VoxelMorph paper \citep{VM}, which is publicly available online. The outputs of this model are a flow field and the corresponding warped moving image. We modified the inputs to this architecture to include the masks for the two input images. We implement SLCC efficiently using sparse convolutional layers \citep{neuron}. Our code is available as part of the VoxelMorph package at \url{http://voxelmorph.mit.edu/}.

\subsection{Experiment Setup}\label{sec:hyperparameter}
 
Since our primary aim is to achieve fast runtimes and high accuracy on \textit{clinical} data, we first evaluate SparseVM and all baselines on the stroke MRI images. 

We also evaluate SparseVM and VoxelMorph on the simulated data enables us to analyze whether our loss function generalizes to anatomical structures other than the ventricles. We also use this data to do a direct comparison of the performance of our sparse version of cross correlation against the usual version of cross correlation. By analyzing the performance of these two learning-based methods, we provide insight into the relative performance of the sparse versions to their isotropic counterparts. 

For the smoothness regularization hyperparameter $\lambda$, we used the value $\lambda=1$ used in the original VoxelMorph papers to guide a parameter search. We then used the validation set to choose an optimal value, which was $\lambda=1.5$. We also did a search over the neighborhood size hyperparameter for SLCC. We provide an analysis below for this value, and report results for SLCC with a neighborhood size $n$=15, which we found to perform well given the sparsity in our data.

\section{Results} 
\begin{figure}[tb!]
\centering
\begin{tabular}{c}
\includegraphics[,width=0.45\textwidth]{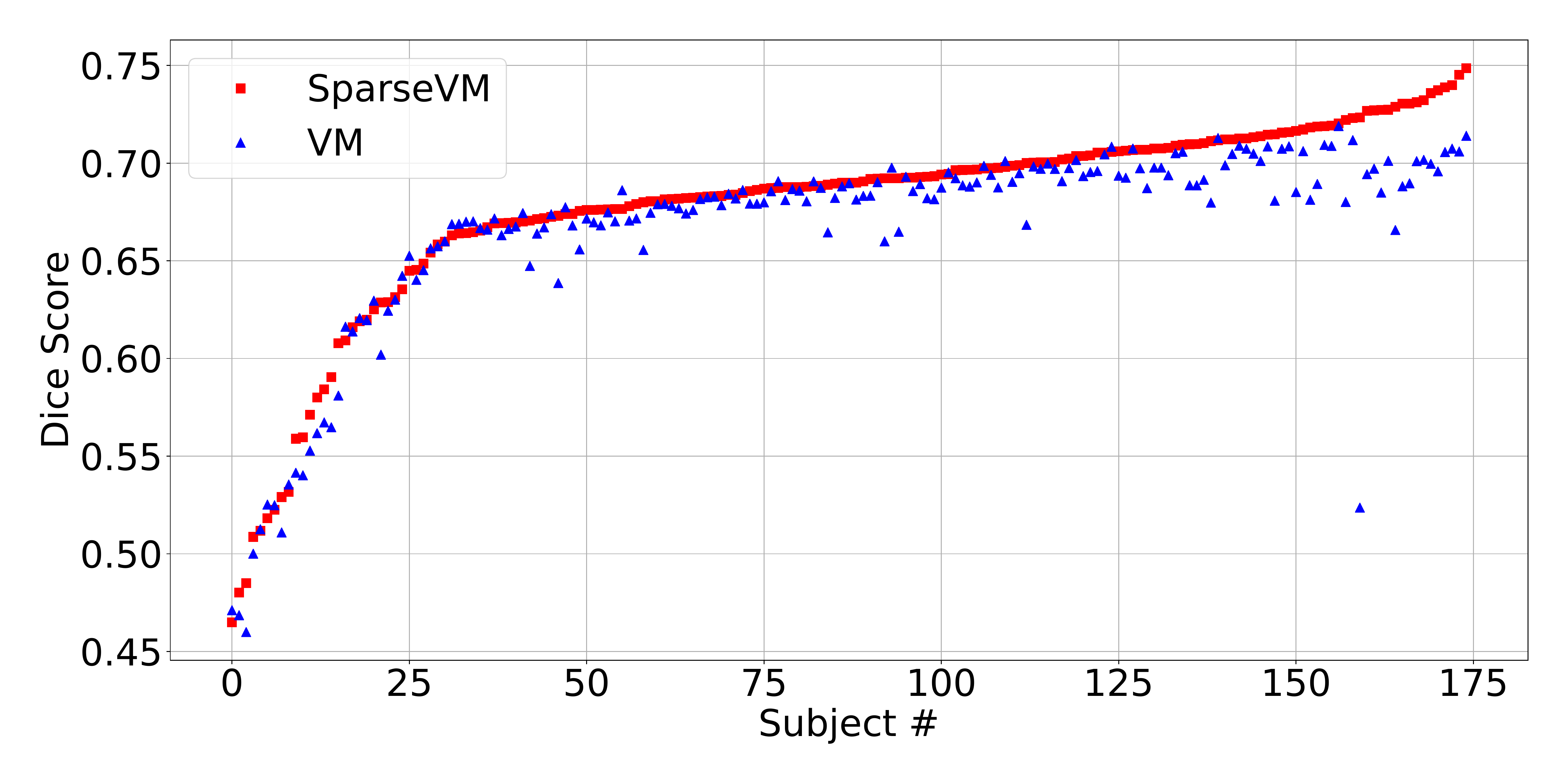} \\ 
\includegraphics[width=0.45\textwidth]{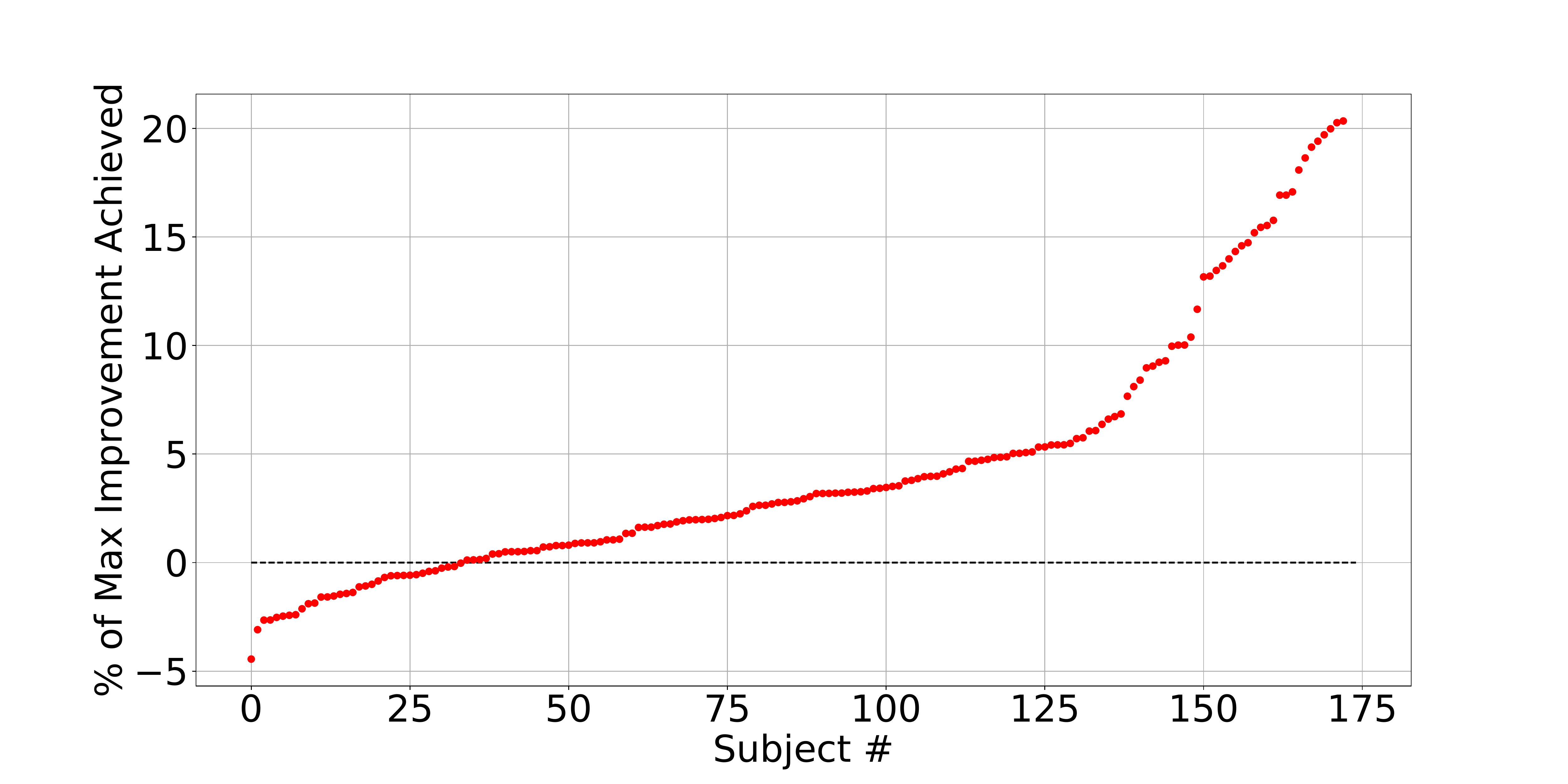}
\end{tabular}
\captionsetup{textfont=normalfont}
\caption{Simulated sparsity results. Per subject average Dice score (in increasing order of the SparseVM Dice score), and percentage of maximum improvement over VoxelMorph achieved by SparseVM. In both, the Dice scores are averaged over all 30 structures. For the bottom graph, a percentage of the maximum possible improvement (in terms of Dice score) was calculated using the average Dice per subject. There are two positive outliers that we exclude from the bottom graph for better visualization, for which SparseVM achieves $27\%$ and $53\%$ of the possible improvement over VoxelMorph.}
\label{fig:diffADNI}
\end{figure}

\begin{table}
\begin{center}
\begin{tabular}{l|c|c}
    \textbf{Method} & \textbf{Average (SD)} & \textbf{Median (MAD)}\\
    \hline
    VoxelMorph LCC & 0.667* (0.052) & 0.682 (0.022) \\
    SparseVM SLCC & \textbf{0.678 (0.053)} & \textbf{0.690 (0.027)}  \\
    \hline
\end{tabular}
\end{center}
\captionsetup{textfont=normalfont}
\caption{Dice scores for VoxelMorph and SparseVM on the simulated sparse ADNI dataset. The average Dice score and median Dice score are computed for all test subjects over 30 different structures. The standard deviations (SD) for the means, and the median absolute deviations (MAD) for the medians are shown in parentheses. We use a * symbol to indicate statistical significance between sparseVM and VM, using a paired t-test with a 0.01 threshold.}
\label{tab:onecolADNI}
\end{table}

\begin{figure*}[tb]
  \includegraphics[width=\linewidth]{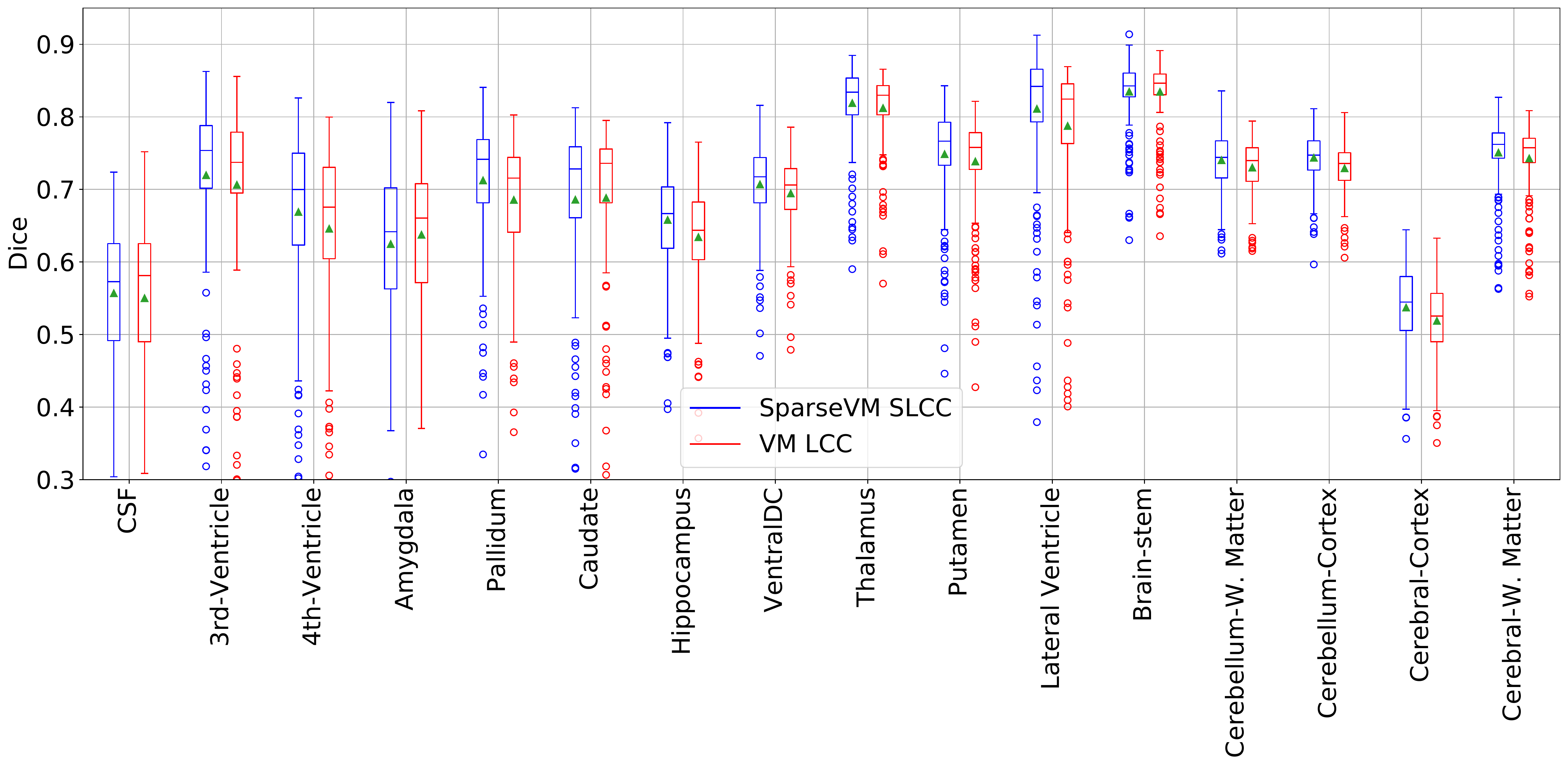}
  \captionsetup{textfont=normalfont}
  \caption{Boxplot of Dice scores for anatomical structures. The x-axis is sorted in increasing order of anatomical structure size. For easier visualization, the left and right side of each anatomical structure (e.g. the left and right lateral ventricles) were averaged together into one boxplot. SparseVM has a higher average for 15 out of the 17 anatomical structures. Fourteen of these differences are statistically significant with a paired t test. The Dice score of the choroid plexus is much lower than all other structures for both methods, and is left out of the boxplot to better visualize the other structures. The average Dice score on this structure was 0.236 for SparseVM and 0.231 for VoxelMorph.}
  \label{boxplotADNI}
\end{figure*}

\subsection{Clinically-acquired Dataset} \label{sec:strokeresults}

% Explain table & mention example segmentation
Table \ref{tab:onecol} summarizes the average and median Dice scores for all test subjects on the ventricles, and the average test runtimes of each method. SparseVM has 1) the highest average and median Dice scores and 2) runtimes on par with VoxelMorph.

Both ANTs and PBR take on the order of two hours on a CPU, while VoxelMorph and SparseVM take less than 10 seconds on a single-threaded CPU and less than half a second on a GPU. To our knowledge, there are no GPU implementations of ANTs or PBR. 

Since the registration results across subjects exhibit high variance, as shown by the standard deviations, we show the Dice score for each subject in Figure \ref{dotplot}. SparseVM using SLCC outperforms the much slower ANTs on 86.7\% of test subjects and the much slower PBR on 69.3\% of test subjects. It has higher accuracy than VoxelMorph on 90.7\% of test subjects. We show two example of subject registration results in Figure \ref{fig:example_segs}. This figure shows a warped atlas segmentation overlaid on a test subject scan for each method. We highlight that these overlaid images are useful for qualitatively showing the accuracy of the resulting flow fields, but the flow fields themselves are the clinically important output. The close-up view in Figure \ref{fig:example_segs} shows that the warped segmentation from our method snaps to edges well. This means that the flow fields produced by our method are accurate near anatomical boundaries. 
%Dice score improvement of SparseVM results compared to VoxelMorph results per subject for the clinical dataset. The data are sorted in increasing order of improvement. 

%\begin{figure}[tb!]
%  \centering
%  \includegraphics[width=\textwidth]{segmentations_nonWMH_labeled.png}
%  \caption{Example results. Warped atlas segmentations overlaid on the subject scan for subjects with different sized ventricles.}
 % \label{WMH_volumes}
%\end{f

%\begin{figure}[tb!]
%  \centering
%  \includegraphics[width=\textwidth]{high_WMH_segs_labeled.png}
 % \caption{Example results. Warped atlas segmentations overlaid on the subject scan for subjects with high WMH burden.}
%  \label{WMH_volumes_SMSE}
%\end{figure}

%Row 3 shows a subject with high white matter hyperintensity (WMH). SparseVM using SLCC registers the subject accurately even though it differs significantly from the atlas and other subject brains.

We find that SparseVM using SLCC tends to perform worse than PBR on subjects for which skull-stripping or affine alignment fails during pre-processing. Figure \ref{bad_vol_SparseVM} shows an example of a test subject that has not been properly skull stripped. Since the atlas does not have a skull and most training subjects are properly skull-stripped, the learning-based methods don't generalize well to outlier subjects that fail skull stripping. The brain anatomy is pulled towards the skull, which suggests the learned models do not properly distinguish the brain from the skull. 

The stroke dataset was compiled as part of a study investigating white matter hyperintensity (WMH) in stroke patients \citep{stroke}. WMH is a pathology, visible on T2 FLAIR scans, that impacts cerebrovascular health. The algorithm used to register images in the population study reported on in \citep{stroke} used flow fields generated by ANTs, a method that we show is much slower and less accurate than ours. Since differences in intensities caused by WMH add to the difficulty of clinical registration, we visually investigate the registration performance of SparseVM near regions of significant WMH. Figure \ref{WMH_volumes} shows registration results for test subjects, illustrating the intensity heterogeneity present in these patients. We observe that SparseVM provides registration that enables the ventricle segmentations to snap well to the anatomical boundaries even when there is high WMH. Consistent with our quantitative results, the close-up regions show that warped atlas segmentations using PBR, VM, and SparseVM outperform the popular toolbox ANTS, with SparseVM (SLCC) performing optimally among these methods. Since SparseVM is both fast and produces more accurate flow fields, its use in clinical studies, such as this one, can reduce errors in propagation of WMH maps and reduce the labor and computational resources needed.

We perform additional analyses on learning-based registration methods to compare the sparse and non-sparse loss functions. SparseVM on average has a Dice score that is approximately 0.02 higher than VoxelMorph's Dice score. The pairwise improvement in accuracy of SparseVM is statistically significant with a p-value of $1.60 \times 10^{-14}$ using a paired t-test. To highlight the magnitude and consistency of this result, Figure \ref{fig:dice_stroke_improvement} shows the Dice score difference as $\frac{\text{SparseVM Dice score - VM Dice score}}{\text{Dice Max - VM Dice score}}$. We use Dice max = 0.9 based on ventricle Dice scores from recent state-of-the-art registration results on isotropic scans \citep{VM,klein}. Since 0.9 is the highest reported Dice, we calculate how much closer SparseVM is able to get to that maximum, in terms of registration Dice, than VoxelMorph.

We also experiment with different neighborhood sizes for SLCC. Figure \ref{neighborhood} shows that choosing the appropriate neighborhood size has an important impact on Dice-score performance. While a neighborhood size of $9^3$ is often used as a good default for isotropic 1mm images, a larger neighborhood size clearly improves performance for clinical image registration for sparse images.

\subsection{Simulated Sparse Dataset}

Table \ref{tab:onecolADNI} shows the average and median Dice scores for both VoxelMorph and SparseVM with the cross correlation loss function. SparseVM outperforms VoxelMorph in both mean and median Dice scores. While the statistics for the stroke dataset shown in Table \ref{tab:onecol} were computed over the ventricle structures, the statistics for the simulated ADNI dataset were computed over 30 different anatomical structures as defined in the FreeSurfer protocol.

In both cases, the sparse cross correlation function outperforms the standard cross correlation function. The average Dice score for the simulated ADNI dataset is lower than that found in the stroke dataset. We hypothesize this is because the simulated dataset averages the Dice score over several smaller anatomical structures in addition to the ventricles. Small mistakes on the boundaries for small structures can drastically decrease the Dice score. 

Since the variance is high between subjects, we also show the Dice score per subject in Figure \ref{fig:diffADNI}. SparseVM performs better than VoxelMorph on 81\% of test subjects and on average has a 0.011 higher Dice score. The improvement is statistically significant when using a paired t-test (p-value is $2.34 \times 10^{-12}$). Figure \ref{fig:diffADNI} also shows the percentage of possible improvement, based on the highest reported Dice score, over VoxelMorph that SparseVM achieves. We show this in order to give intuition about the difference in Dice score and the Dice score range. Figure \ref{fig:diffADNI} follows a trend similar to Figure \ref{fig:dice_stroke_improvement} for the stroke data.

Figure \ref{boxplotADNI} shows a boxplot of the average Dice score per label for SparseVM and VoxelMorph. SparseVM has a higher average for 15 of the 17 structures. Eleven of these improvements are statistically significant with p-values $< 2\times 10^{-6}$ with a paired t test and three of these improvements are statistically significant with p-values $< 0.02$. VoxelMorph performs better on two of the smallest structures: the amgydala (statistically significant with p-value $5.95\times10^{-10}$) and caudate.

\section{Conclusion}
We introduced SparseVM, the first method to use neural networks to register sparse, low-resolution clinical scans. SparseVM is over 1000$\times$ faster than classical methods that achieve similar accuracy for clinical image registration. On clinical images, SparseVM requires about the same runtime as the best learning-based method for registering scans, while also providing a statistically significant improvement in accuracy. SparseVM can accurately register pairs of clinical scans in under a second on the GPU, thus enabling clinical image analyses that were not previously feasible.

\begin{acks}
This project was funded in part by the Wistron Corporation and NIH grant 1R21AG050122.
\end{acks}

%\bibliographystyle{ACM-Reference-Format}
%\bibliography{CHIL}
\input{CHIL.bbl}

\end{document}